# The Common Core Ontologies

Mark JENSEN [a,b,1], Giacomo DE COLLE [b,c], Sean KINDYA [c],
Cameron MORE [b,d], Alexander P. COX [b,d], and John BEVERLEY [b,c,e]

[a] *U.S. Customs and Border Protection*
[b] *National Center for Ontological Research*
[c] *University at Buffalo*
[d] *CUBRC, Inc.*
[e] *Institute for Artificial Intelligence and Data Science*

**Abstract.** The Common Core Ontologies (CCO) are designed as a mid-level ontology suite that extends the Basic Formal Ontology. In 2017, CUBRC, Inc. made CCO openly available. CCO has since been increasingly adopted by a broad group of users and applications and is proposed as the first standard mid-level ontology. Despite these successes, documentation of the contents and design patterns of the CCO has been comparatively minimal. This paper is a step toward providing enhanced documentation for the mid-level ontology suite through a discussion of the contents of the eleven ontologies that collectively comprise the Common Core Ontology suite.

## 1. Introduction

Ontologies can be understood along levels of generality. Top-level ontologies represent entities at the highest level of generality, reflected in terms such as "part of" and "role" [1]. Basic Formal Ontology (BFO) [2] is an example of such an ontology. Domain ontologies are designed to represent entities within specific areas of interest and so contain more fine-grained terms, such as "variegated Ficus" or "muscarinic acetylcholine receptor". The Occupation Ontology (OccO) is an example of a domain ontology, as it is specifically designed to represent international occupation codes [3]. Mid-level ontologies fall somewhere between top-level and domain ontologies, as they are designed to represent entities at a lower level of generality than a top-level ontology, but a higher level of generality than any domain ontology. The Common Core Ontologies (CCO) suite [4, 5] is an example of eleven ontologies which, collectively, comprise a mid-level ontology.

CCO – initiated by CUBRC, Inc. in 2010 under an IARPA Knowledge Discovery and Dissemination grant – is widely used in defense and intelligence sectors to support data standardization, interoperability, reproducibility, and automated reasoning across numerous domains [6, 7, 8, 9, 10]. Accordingly, CCO development and application was, for many years, conducted without much transparency. As of 2017, however, CCO has been available under a BSD-3 license with a public GitHub repository open to collaboration. Making CCO publicly available has led to significant increase of interest in CCO development. For example, in 2022 the Institute of Electrical and Electronics Engineers (IEEE) P3195 Standard for Requirements for a Mid-Level Ontology and Extensions working group [11] initiated a review of CCO to become the first mid-level ontology standard. More recently, in 2024 CCO was endorsed as a "baseline standard" for formal ontology development across the Department of Defense and Intelligence Community [12].

---
[1] Corresponding Author: Mark Jensen, mpjensen@buffalo.edu

CCO documentation consists of user guides [4, 5] describing how to apply CCO design patterns to specific use cases, but at present there are no substantial explanations of and motivations for ontological commitments adopted by CCO. We will begin to address this gap in what follows by providing an overview of several important features of CCO. More specifically, we outline and defend the methodological and ontological commitments underwriting CCO development, its fundamental structure, complexities of a handful of design patterns, and limitations of future directions for the ontology suite.

## 2. The Common Core Ontologies

CCO was developed and is maintained in compliance with Open Biological and Biomedical Ontologies (OBO) Foundry standards, though it is not part of the foundry [9] owing to its scope. Principles of the OBO Foundry include, for example, requiring ontologies be constructed using an unambiguous syntax, include unique identifiers, unique definitions, and be publicly available [13]. Moreover, as with most terms found in OBO Foundry ontologies, every term in CCO ultimately extends from some term found in BFO [1, 5, 6]. To appreciate CCO, a brief discussion of BFO is in order.

*2.1 Basic Formal Ontology*

Each of the eleven ontologies comprising CCO extends from BFO and so inherits its methodological commitments [6, 14]:
- *Realism* – Ontologies should aim to describe reality, rather than language, concepts, or mental representations about reality.
- *Fallibilism* – Subject-matter experts provide our best evidence for the nature of reality, but because these sources are fallible, ontologies are as well.
- *Adequatism* – Ontologies should represent all entities in a domain taken seriously by subject-matter experts and resist paraphrasing away talk of entities in terms of others.

Regarding *Realism*: Because language, concepts, mental representation, and so on exist in reality, each is also within scope of CCO. Regarding *Fallibilism*: CCO has undergone five substantial updates, reflected in releases which can be found in its GitHub repository. Each release emerged out of updates based on stakeholder needs and corrections. Regarding *Adequatism*: CCO is designed to provide developers working with domain ontologies definitions, notes, and labels, that more closely resemble natural language than what is found in BFO, which is a significant benefit given BFO's sometimes obscure commitments. In addition to these commitments, CCO adopts the "middle-in" ontology development strategy [15]: bottom-up ontology development within the constraints of a top-level ontology.

CCO adopts BFO's division between **occurrent**[2] and **continuant**, as well as subclasses and relationships among them. **Occurrents** are extended over time and have temporal parts, such as eating or walking, each of which is an example of the **process** subclass of **occurrent**. **Temporal region** represents, in effect, when **processes** unfold, including **process boundaries**, the beginnings and endings of **processes**.

Instances of **continuant** lack temporal parts, endure through time, and participate in instances of **occurrent**. **Independent continuants** are **continuants** that do not depend on anything for their existence [1]. For example, some shape and mass depend on a given baseball for their existence, but not the other way around. **Independent continuant** is further divided into **material entity** and **immaterial entity**, the former having material parts and the latter lacking them.

The mass and shape of a baseball are **specifically dependent continuants**, instances of which always depend for their existence on an **independent continuant**. More specifically, they are **qualities** since they manifest fully whenever they exist at all. In contrast, **realizable entities** need not fully manifest at any time they exist. For example, the solubility of a portion of salt is a **disposition** which may or may not manifest. **Dispositions** are internally grounded, meaning that for a **disposition** to begin or cease to be borne, its bearer must undergo a physical change. For example, the solubility of salt is a **disposition** and a portion of salt losing its solubility must undergo some physical change to its lattice structure. In contrast, a **role** is a **realizable entity** that is externally grounded, meaning that the gain or loss of this **disposition** does not necessarily affect the physical makeup of its bearer. For example, being a student is a **role** borne by a person enrolled, say, at a university to study physics [14]. The class **disposition** itself has a single subclass: **function**, the bearing of which are the reasons their bearers exist, e.g. a toaster exists to toast bread, a heart exists to pump blood.

**Generically dependent continuant** is the sibling class of **independent continuant** and **specifically dependent continuant**, where one finds entities that may be copied across bearers. For example, your computer monitor could bear any of the following distinct patterns: 'π', 'pi', '3.14...', or '3.14159265358979323...', and all of these would convey the same information. Similarly, an iPad screen might exhibit smaller patterns which also convey the same information.

*2.2 Common Core Overview*

CCO modules are not intended to provide an exhaustive taxonomy of all possible entities within scope but instead provide guardrails for extending the top-level semantics of BFO into more specific domains. For example, the Artifact Ontology is a guide for extending the semantics of **artifact** into specific domains of interest, such as medical artifacts, scientific artifacts, military artifacts, and so on. Such guardrails promote data FAIR-ness [18] and datasets accurately tagged with CCO terms are natively interoperable with other projects that adopt CCO, promoting extensibility.

Because CCO is designed with such flexibility in mind, CCO terms and relations do not exhibit extensive axiom constraints. For example, there is no axiom connecting the class **person** to the class **weight** – a **quality** inhering in a **material entity** in virtue of its location in a gravitational field. This allows users to model with CCO **person** without being forced to represent **weight** if they have no need to do so. Users are, of course, welcome to add axiom constraints as needed.

---

[2]In the sequel, **bold** will be used to represent classes, *italics* to represent relations.

**Table 1** provides a list of the eleven modules of the CCO suite, alongside corresponding scope statements [17].

*Table 1. List of Common Core Ontologies and Scope Statements*

| Common Core Module | Scope |
|---|---|
| *Geospatial Ontology* | Designed to represent sites, spatial regions, and other entities, especially those that are located near the surface of Earth, as well as the relations that hold between them. |
| *Information Entity Ontology* | Designed to represent generic types of information as well as the relationships between information and other entities |
| *Event Ontology* | Designed to represent processual entities, especially those performed by agents, that occur within multiple domains. |
| *Time Ontology* | Designed to represent temporal regions and the relations that hold between them. |
| *Agent Ontology* | Designed to represent agents, especially persons and organizations, and their roles. |
| *Quality Ontology* | Designed to represent a range of attributes of entities especially qualities, realizable entities, and process profiles. |
| *Units of Measure Ontology* | Designed to represent standard measurement units that are used when measuring various attributes of entities. |
| *Currency Unit Ontology* | Designed to represent currencies that are issued and used by countries. |
| *Facility Ontology* | Designed to represent buildings and campuses that are designed to serve some specific purpose, and which are common to multiple domains. |
| *Artifact Ontology* | Designed to represent artifacts that are common to multiple domains along with their models, specifications, and functions. |
| *Extended Relations Ontology[3]* | Designed to represent many of the relations that hold between entities at the level of the mid-level Common Core Ontologies. |

Terms within a given CCO module need not all extend from a single BFO class or relation. For example, the Quality Ontology classes **magnetism** and **mass** extend from **realizable entity** and **specifically dependent continuant**, respectively. Moreover, though the CCO ontologies have been modularized for ease of use, they are in many ways designed to be used together, as many CCO modules import terms from others. For example, the Information Entity Ontology imports the Time Ontology and Geospatial Ontology, while the Agent Ontology imports the Information Entity Ontology. Consequently, our exposition of the CCO suite will be driven by design patterns borne out of applications of these ontologies to address real-world use cases.[4]

CCO modules are checked for logical consistency using the Hermit reasoner [19] following each release. A merge of the modules is similarly checked for consistency.

## 2.3 Tracing through Space and Time

CCO has been used to integrate heterogenous data concerning entities that move through space over time. By introducing entities such as **geospatial boundary**, **geospatial line**, and **geospatial polygon**, CCO's geospatial design patterns resemble those of GeoSPARQL which maintains a vocabulary of points and polygons from which users may construct query patterns concerning spatial location [20].[5] However, GeoSPARQL does not provide query extensions or native support for interactions between both space *and* time, instead directing users to leverage OWL TIME to model the latter [22]. In contrast, CCO integrates spatial, temporal, and spatiotemporal aspects of tracking.

---
[3]The Extended Relations Ontology is based on the Relations Ontology maintained by the Open Biological and Biomedical Ontologies Foundry [17].

[4]Owing to space constraints, we do not discuss the Currency Unit Ontology, which extends the Units of Measurement Ontology. This ontology treats currency as a measurement of financial value of some entity which may differ based on the currency, e.g. $1 USD, $1.35 CAD and so on.

Suppose there is a need to represent the path taken by a ground vehicle over some geospatial region. **Figure 1** illustrates how such a use case would be modeled using CCO. The class **material artifact** - a **material entity** designed by some agent to realize some **function** - is the parent of **vehicle**, instances of which convey **material entities** from one location to another. Subclasses of **vehicle** are divided largely along the lines of **aircraft**, **ground vehicle**, **spacecraft**, and **watercraft**, all of which are designed to realize some conveyance **function** across some environment type [23]. While any instance of **truck** is a **ground vehicle**, the latter class is further divided into **rail transport vehicle** – conveyance by railway - and **ground motor vehicle** – conveyance by motive power by an engine absent rails – where we find **truck**.

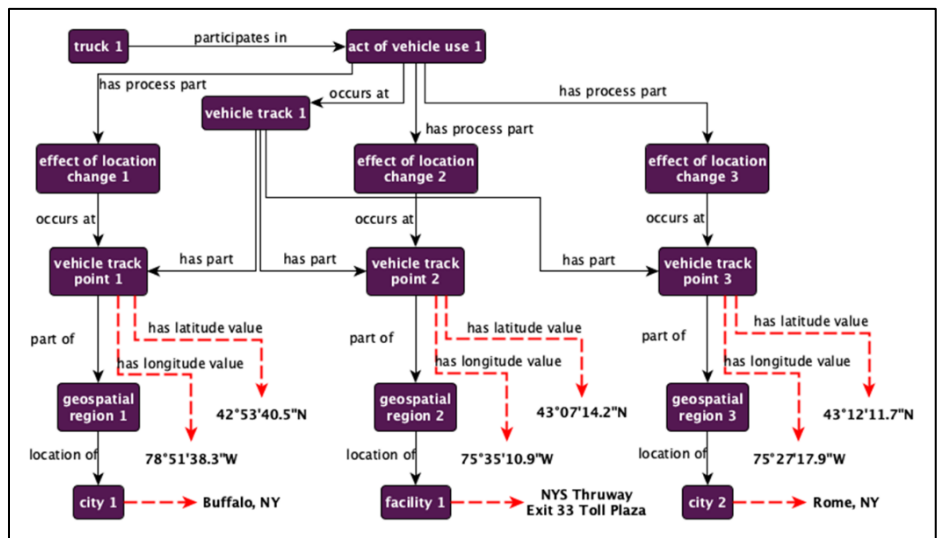

*Figure 1: Tracing a Truck across Geospatial Regions*

Tracing the path of an instance of **truck** involves identifying an **act of vehicle use** in which that **truck** *participates*. The **act of vehicle use** in turn *has process part* some location changes which *occur at* instances of **vehicle track point**. Generally, the relation *occurs at* holds between a **process** and a **site**, where a **site** is a three-dimensional **immaterial entity** whose boundaries coincide with some **material entity**, e.g. a hole in a straw, the trunk of a car. To say then that a process part of the **act of vehicle use** *occurs at* some **vehicle track point** is to imply the latter is a **site**. Each instance of **vehicle track point** is associated with latitude and longitude text values, and each is a *part of* distinct instances of **geospatial region**, which is a **site** at or near the surface of the Earth. Each geospatial region has a different **location**, such as Buffalo NY, or the New York State Thruway Exit 33 Toll Plaza, or Rome NY.[6]

---
[5]The recent release of GeoSPARQL provides a preliminary mapping into BFO [21], and so a foundation on which to align GeoSPARQL and CCO.

The temporal aspect of this scenario is illustrated in **Figure 2**. Suppose an **act of vehicle use** *occurs at* the Baghdad **city** of al-Kadhimya over the course of several months. One such occurrence was observed happening at a time of day during the month of May 2004. Any **act of vehicle use**, like all processes, occurs on a **temporal interval**, which is a continuous temporal region of one-dimension exhibiting no gaps, such as a portion of a particular day. While the observation is stamped to a particular temporal instant, we know that time of day it is part of larger interval, say, the month of May, which itself is an interval during Spring 2004, which in this scenario is an interval during a **multi-month temporal interval**. Note that, despite the label choice, **multi-month temporal interval** may, unlike the BFO class **temporal interval**, exhibit discontinuities and temporal gaps. The class exists to track repeated occurrences of a **process** across months. For more fine-grained representations, CCO additionally contains similarly contains **multi-day temporal interval**, **multi-second temporal interval** and so on.

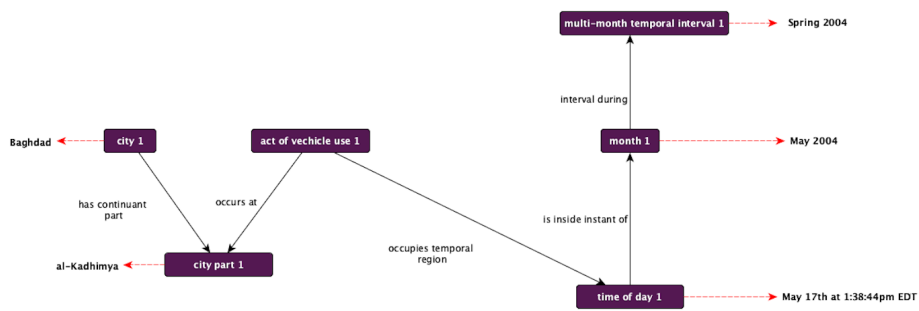

*Figure 2*: Tracking Processes across Temporal and Spatial Granularities

These resources allow for the representation of partial descriptions of **processes** that refer to the same event using different granularities of time. For example, the statements "The truck was in Baghdad at 13:38EDT on May 17[th], 2004" and "A truck was in al-Kadhimya the afternoon of March 17[th], 2004" may refer to the same event, and if so, would be linked using *interval during* or *is inside instant of* relations.

2.4 The Aboutness of Information

The Information Entity Ontology distinguishes content of information both from the **information bearing entities** which may carry that content and from the patterns exhibited by those **information bearing entities**. A computer monitor screen, for example, bears **qualities** such as shape and color that are said to *concretize* **information content entities**, i.e. **generically dependent continuants** that are *about* something. These distinctions allow for flexible representations of various relationships arising among **information bearing entities**, **patterns**, and **information content entities**. Any of the following patterns 'π', 'pi', '3.14...', or '3.14159265358979323...' on your monitor might *concretize* the same **information content entity**. Similarly, such patterns in a textbook would *concretize* the same **information content entity**.

---

[6]The CCO class **city** is found in the Agent Ontology while New York State Thruway Exit 33 Toll Plaza is found in the Facility Ontology. While any **facility** is an **artifact**, the number of subclasses of the former in CCO warranted creating the Facility Ontology as an extension of the Artifact Ontology.

Importantly, in making this trifold distinction, CCO denies that information transmission, provenance, and evaluation can be adequately represented without reference to information carriers. Carriers are, for example, crucial when modeling the provenance and pedigree of data across multi-modal sensors [24]. That said, users are not required to track provenance; CCO includes an annotation property *is tokenized by* to link literal values directly to instances of **information content entity** without having to represent relevant **information content entities**.

Subclasses of CCO **information content entity** are inspired by literature on speech act theory [25]. For example, **designative information content entities** are used to uniquely denote entities, while **directive information content entities** consist of either propositions or images used to prescribe behaviors, actions, designs, etc. The class **descriptive information content entity** consists of propositions used to describe some entity and is the parent to the extensive collection of measurement and measurement unit CCO classes. CCO distinguishes what is being measured, information about what is measured, units encoding measurements, and the findings regarding measurements. For example, measuring John's height in inches involves John, a length **quality** that *inheres in* John, the inch unit of measure, and the value associated with John's height, e.g. "70".

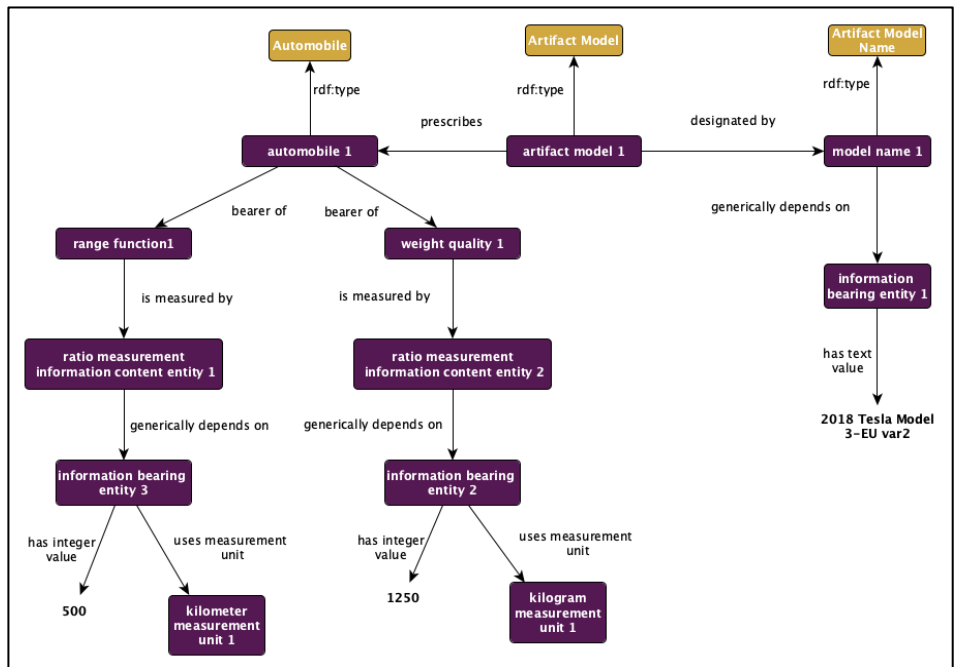

*Figure 3*: Measuring Outcomes of an Artifact Model Prescription

To illustrate features of CCO's information design pattern, consider that automobiles are often designed by agents or organizations according to some blueprint or modeling pattern. The Agent Ontology contains classes for **agents**, **organizations**, and **roles** borne by either. In CCO, **agents** are in every case those **material entities** capable of performing **planned acts**, i.e. acts directed by some **directive information**

**content entity**. Moreover, by leveraging a sub-relation of *participates in*, namely *agent in*, CCO distinguishes between **agents** making causally relevant contributions to some **process** as opposed to passive contributions. For example, engineers working for Honda at some point created a blueprint for the Honda Civic, and so provided causally relevant contributions to the creation of this blueprint.[7] Expanding on this example, CCO introduces the class **artifact model**, a **directive information content entity** that prescribes a common set of **functions** and **qualities** to *inhere in* a set of artifact instances. Instances of **artifact model** are, moreover, designated by specific **artifact model names**, such as "2018 Tesla Model 3-EU var2" which may similarly be represented using CCO resources.

Given an **artifact model** prescribing the production features for a type of automobile, manufacturers also engage in **planned acts** the goal of which is to satisfy the **artifact model** prescriptions. Manufacturers may, for example, produce an automobile that is meant to bear a certain **weight** or have a certain **transportation range function**. These dependent entities will then inhere in the automobile produced and may be measured by some **information content entity**. As depicted in **Figure 3**, each **ratio measurement information content entity** corresponds to its own **information bearing entity**, which we may assume without loss of generality that in this case is part of some database or table. With respect to the **weight** of the automobile, the relevant database part has a literal value "1250" and *uses measurement unit* **kilogram measurement unit**.

**Artifact models** are rarely, if ever, faithfully produced. CCO resources allow for representations of the goals of an **artifact model**, the extent to which attempts to produce that model were successful, and manufacturer plans in pursuit of such production. When representing data reflecting failures of plans, missed opportunities, or perhaps even unobtainable goals, such nuanced representations are invaluable.

*2.5 Change and Constancy over Time*

Because ontologies are typically represented in the Web Ontology Language [26], object properties found in implementations of CCO are restricted to being binary. This constraint results in modeling challenges when, say, one needs to represent change over time, e.g. John has an arm part at time 1 but not at time 2. During the ISO/IEC standardization of BFO, a strategy was devised to address this challenge that relied on the temporal quantification of object properties. For example, BFO *continuant part of* was split into *continuant part of at all times* and *continuant part of at some time* [2]. Following BFO's approval as an ISO/IEC standard, CCO adopted this strategy. However, difficulties in deploying temporally quantified object properties in applications of BFO led to the creation of a "core" OWL version of BFO – which adopts object properties without temporal qualification such as *continuant part of* - and a separate "temporalized" OWL version of BFO – which contains temporally qualified object properties such as *located in at some time* and *member part of at all times* [27]. The most recent release of CCO adopts only the core object properties of BFO, dropping object properties such as *member part of at all times*.

---

[7] Our tracking example from the previous section could be embellished in the following manner: Instances of **act of vehicle use** are **planned acts** and so alongside the instance of **truck**, some **agent** may also *participate in* that act i.e., by driving the **truck**. A passenger passively riding in the **truck** would not be an *agent in* this act.

As illustrated in the tracing design pattern above, CCO provides adequate resources for representing time without delving into the more sophisticated temporal representations of previous versions of BFO. Nevertheless, there is a need to represent constancy over time, such as Mary's temperature being normal over a series of measurements, the average altitude of flight, or how long a suspect was detained. Consider repeated measurements of Mary's temperature over the course of a day. One strategy for representing the phenomenon in question is to assert that Mary is the bearer of an instance of **temperature** that carries some text value within a normal range, and so warrants classification as a 'normal measurement'. Such a strategy does not, however, provide the ability to represent a single measurement datum in a record indicating that, say, "On Friday 3/15/24, Mary's temperature was normal."

The strategy adopted by CCO is to introduce the class **stasis**, a **process** in which an aspect of one or more **independent continuants** endures in an unchanging condition. Intuitively, the constancy of Mary's 'normal' temperature is represented by connecting measurements of Mary's **temperature** to a proper *process part* of bodily processes she *participates in* over some interval, during which **qualities** that impact Mary's **temperature** are measured as within some range of normalcy. In this case, the relevant proper *process part* count as a **stasis**. Put another way, we can think of Mary's 'normal' **temperature** as *the subject of* some **interval measurement information content entity** - since such measurements typically have no absolute zero. Mary's **temperature stasis** *has participant* the **temperature** that is *subject of* that **interval measurement information content entity**. Mary's 'normal' temperature is then captured by asserting Mary's **temperature stasis** *occupies temporal region* for some **temporal interval** designated by a **date identifier**, a CCO class used to denote specific days, which in this case will have value "3/15/24".

This pattern is designed to be general, e.g. applying to the average altitude of a jet, location of a detainee, and so on. **Figure 4** illustrates the design pattern via Mary's temperature.

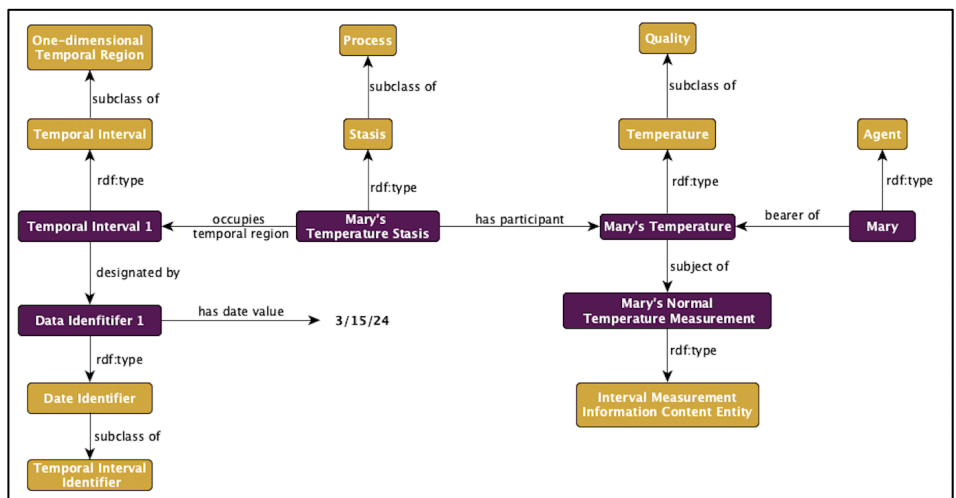

*Figure 4: Measuring Constancy over Time*

Related, CCO provides resources for the representation of the gain or loss of dependent entities because of some **process**. The extensive class structure of CCO's

Quality Ontology – initiated by adapting classes from the Phenotypic Trait Ontology (PATO) [28] - provides a storehouse of **qualities** that may be involved in such processes. The CCO class **change** represents a **process** in which some **independent continuant** endures and 1) one or more of the dependent entities it bears increase or decrease in intensity, 2) the entity begins to bear some dependent entity or 3) the entity ceases to bear some dependent entity. Following the BFO hierarchy of dependent entities, subclasses cover decrease, gain of, increase of, and loss of dependent continuants, each of which contains subclasses for **specifically dependent continuants** and **generically dependent continuants**. Regarding the former, if a portion of $H_2O$ is frozen, it loses its liquidity **disposition**, represented as this portion of $H_2O$ participating in a **loss of disposition**. Regarding the latter, receipt on your local network of a PDF file sent via email reflects a **gain of generically dependent continuant**.

### 3. The Future of Common Core Ontologies

While the CCO has adhered to best practices set forth by the OBO Foundry and BFO developers, it has also been charting a novel set of representational challenges. Early CCO developers had little access to robust BFO-conformant ontologies to reuse. After over a decade of development, CCO is continually undergoing refinement. Much of this exists at the lower levels, often directed by mission-specific applications, where users make term requests noting gaps or points of ambiguity in definitions. Nonetheless, there also exist more substantive areas where users have noted a lack of clarity or outright badly formed semantics in CCO.

One such area in need of improvement is the representation of **stasis**. There have been several lengthy discussions on the CCO repository issue tracker [29, 30] about, say, whether **stasis** is proper subclass of **process**, as well as, perhaps more importantly, the intended usage and semantic patterns associated with data about **stases**. Presently, there is no clearly articulated pattern for what *participates in* a **stasis**. It is not enough to simply assert that Mary's **temperature** is *participating in* the **stasis** without also including the measurement about it. Mary bears her temperature for the extent of her life, and so, plausibly it will have many such instances of **information content entity** that describe it. It is, however, unclear which are associated with a particular **stasis**. The design pattern described above demonstrates the importance of making explicit what remains stable during **stasis**.

Additionally, there is a need for explicitly aligning CCO to existing ontologies or other terminological standards. As of this submission, there has been no overt effort by CCO developers to publish vetted mappings to other ontologies. As the user base of CCO grows, it makes sense for CCO to explicitly endorse compatibility with other ontologies. Both the OBO Foundry and IOF are developing core mid-level ontologies to aid in unifying their various extension modules. With the increasing uptake of enterprise semantic architectures and foundry-style ecosystems, these mappings would improve interoperability across with nearby efforts. The design patterns represented in this document are rather complex modeling schemas, but within them exist smaller chunks of repeatable and very commonly reused design patterns for simple datums, such as vehicle weight, or person name, or geospatial identifiers.

CCO's representation of measurement units is more exhaustive and semantically rich than most other domain ontologies. However, many such standards exist for incorporating measurement units. It is not clear many users of CCO want or see benefit to inheriting such a complex representation of measurement units. In the vein of offering mappings to existing standards described above, CCO could address the mismatch and lack of consistency between how it represents measurement units, so that users are able to easily convert data across ontology standards. For example, the Quantities, Units, Dimensions and Types (QUDT) Ontologies are commonly used across graph-based ontology applications [31]. QUDT contains a small number of primitive classes including quantity value, quantity kind, unit, and quantity kind dimension vector. QUDT's primary use case is the conversion of one unit type of measurement into another unit type, not about linking data to things in the real world, the latter being the focus of CCO. Nevertheless, mapping between CCO and QUDT would allow users to take advantage of the rich measurement vocabulary offered by CCO as well as the inter-unit conversion SPARQL queries that underwrite the QUDT data model.

## 4. Conclusion

CCO is a mid-level ontology suite that extends the Basic Formal Ontology. CCO has since been increasingly adopted by a broad group of users and applications and is currently being vetted to become the first standard mid-level ontology. Despite these successes, documentation of the contents and design patterns of the CCO has been comparatively minimal. This paper is a step toward providing enhanced documentation for the mid-level ontology suite through a discussion of the contents of the eleven ontologies that collectively comprise the Common Core Ontology suite.

## 5. Code Availability

CCO ontologies are available in .ttl files on the CCO GitHub repository (https://github.com/CommonCoreOntology/CommonCoreOntologies); the repository also includes a .ttl file which is a merge of the eleven CCO ontologies and official documentation. CCO is maintained under the 3-Clause BSD-3 License (https://opensource.org/license/BSD-3-clause) which allows for the copying, redistribution, and adaption of the ontology for any purpose, with the caveat that the trademark owner is not assumed to endorse such use unless giving explicit permission.

## 6. Acknowledgements

Many thanks are owed to the large and growing body of contributors to CCO over the years, with particular gratitude owed to Ron Rudnicki, Barry Smith, Alan Ruttenberg, Victor Rohr, and James Schoening for initiating and nourishing CCO since its inception. Additionally, thanks to the *IEEE P3195 Mid-Level Ontology and Extensions Working Group* for continued work towards making CCO an international standard: Brian Haugh, Neil Otte, Austin Liebers, Carter Benson, Alec Sculley, David Lutz, Alan Belasco, Forrest Hare, Eric Merrell, Tim Prudhomme, Jonathan Vajda, and Steven Wartik.